# A Real-time Anomaly Detection Using Convolutional Autoencoder with Dynamic Threshold

Sarit Maitra
*Alliance School of Business*
*Alliance University*
Bengaluru, India
sarit.maitra@gmail.com

Sukanya Kundu
*Alliance School of Business*
*Alliance University*
Bengaluru, India
sukanya.kundu@alliance.edu.in

Aishwarya Shankar
*Alliance School of Business*
*Alliance University*
Bengaluru, India
saishwaryamba722@bus.alliance.edu.in

*Abstract*— **The majority of modern consumer-level energy is generated by real-time smart metering systems. These frequently contain anomalies, which prevent reliable estimates of the series' evolution. This work introduces a hybrid modeling approach combining statistics and a Convolutional Autoencoder with a dynamic threshold. The threshold is determined based on Mahalanobis distance and moving averages. It has been tested using real-life energy consumption data collected from smart metering systems. The solution includes a real-time, meter-level anomaly detection system that connects to an advanced monitoring system. This makes a substantial contribution by detecting unusual data movements and delivering an early warning. Early detection and subsequent troubleshooting can financially benefit organizations and consumers and prevent disasters from occurring.**

## I. INTRODUCTION

OECD[1] predicts the world economy would require 80% more energy in 2050 without a new measurement policy. Global power consumption is rising at an alarming rate, and buildings (residential and commercial) account for around 40-45% of global consumption ([1]; [2]; [3]; [4]; [5]). According to IEA[2], while the global growth in electricity demand eased slightly to 2.2% in 2023, it is projected to accelerate to an average of 3.4% from 2024 through 2026. Because of the pervasive misuse of residential power consumption behaviors and controls, it is estimated that 15–30% of the energy utilized during building operations is lost owing to malfunctioning equipment, poor operation protocol, and poor construction design ([6]; [3]). The frequent occurrence of anomalies captured by real-time smart metering systems attributed to a combination of technical, environmental, human, and operational factors ([7]; [8]; [9], [10]).

Anomalies refer to data points that significantly deviate from the rest of the dataset. These are the minority datasets, but they have the potential to lead to serious implications if not immediately identified and cured. A clear mathematical definition cannot explain anomalies since their detection is a more subjective decision than a routine assessment. Therefore, a real-life detection method prefers to look for a real-time detection system matching the semantics of the specific application context. Our work investigates the data collected from smart metering systems to develop a real-time anomaly detection system. Table I presents the identified research gap vis-à-vis the key contribution of our work.

TABLE I. RESEARCH GAP AND KEY CONTRIBTUION.

| Research gap | Our contribution |
|---|---|
| Research on building energy consumption is not adequate; however, it is showing rapid progress. The number of publications has grown significantly (Han and Wei, 2021). The field is dominated by academic scholars and lacks the perspective of industry practitioners. Though the existing researchers covered diverse fields, data-driven anomaly detection using multi-variate analysis in this domain is not prevalent in real life. Most of the existing work used normal distributions for anomaly classification (see Liu and Nielson, 2018). Though the statistical method is effective, it must be used with other advanced techniques for better efficiency. | We present a combination of advanced data mining and machine-learning (ML)-based techniques. It combines statistical and unsupervised convolutional autoencoders to provide a near-real-time anomaly detection system. The work is tested with real-life data obtained from a smart metering system. We propose a dynamic thresholding approach based on moving averages and Mahalanobis distance (MD) to improve the system's accuracy, responsiveness, and adaptability and make it suitable for tracking aberrant patterns in real-world situations. Our proposed system presents a scalable and near-real-time anomaly detection mechanism. |

In the past, researchers applied ML for anomaly detection in power consumption (see [11]; [12]). However, their studies pertain to fraud consumption and do not provide any real-time reporting. Moreover, though scalable anomaly detection for smart meters has been investigated (see [13]), their work has not been evaluated to detect a wide range of anomalies, including missing values, negative energy consumption, device errors.

The deployment of smart meters in smart cities demonstrates paradigm shift in energy management [7]. Smart meters are digital devices that measure and record consumption in real-time. These devices are rapidly becoming popular in modern energy management and have opened avenues for data analysis. Hence, research on anomaly detection for power consumption has gained momentum in recent times ([14]; [15]; [16]). Researchers found that efficient anomaly detection is one of the best ways to reduce waste during building operations [3]. The availability of real-time data and increased research interest

---

[1] https://shorturl.at/dBEX0
[2] https://shorturl.at/rGKN6

have turned anomaly detection into sophisticated data mining work.

Researchers typically classify anomalies into three categories: point, contextual, and collective anomalies. Our focus here is on contextual and point anomalies. An early alert would help reduce the anomalies and prevent financial losses. Various research works have reported about the potential and possibility of employing artificial intelligence (AI) to detect anomalous patterns ([17]; [15]). Despite the potential of AI, real-time energy data does not contain annotations or labels that specify when anomalies occur [18]. Thus, unsupervised ML approaches are favored in this dataset. However, it comes with its own set of challenges, which may include appropriate feature representations, selecting suitable algorithms, addressing data noise and variability, and interpreting detected anomalies.

### A. Proposed approach and applicability

A convolutional autoencoder offers two primary benefits. Firstly, since the network is able to generate discriminative features on its own, feature engineering is not required. Second, since the process is unsupervised, by training an autoencoder on normal energy consumption patterns, the model can learn to identify deviations without the need for labeled data. The dynamic threshold adjusts over time to the changes in the data distribution, allowing the system to adapt to varying conditions and maintain sensitivity to anomalies. The approach can effectively distinguish between normal variations and true anomalies. Moreover, we emphasize the importance of visualization techniques for interpreting detected anomalies and providing human-readable insights. We integrate additional data sources, e.g., holiday schedules and weather data, to enrich the analysis and improve its accuracy. By incorporating contextual information into the process, the system can better capture the complex dependencies and external factors that influence consumption patterns.

## II. DATA DESCRIPTION & ANALYSIS

This data presents hourly energy measurements of three factory monitoring systems published by Schneider Electric[3]. and spans 2012–2016. It is supported with three additional datasets: (1) metadata offering location information and general descriptive features of the meters; (2) a holiday dataset summarizing the public holidays; and (3) corresponding hourly weather data.

The hardware specifications used in this work are Core (TM) i5-4570 CPU at 3.20GHz, 16 GB of DDR4 SDRAM, Windows 11 64-bit, and Nvidia GeForce RTX 2070. The software description is Python v3.10.12, Jupyter Notebook v5.3.0, Pandas v1.5.3, NumPy v1.23.5, Matplotlib v3.6.3, Google Cloud.

For energy-related data, we combined the data sets on the meter ID; for weather and holiday data, we did the same with the date stamp. We focus on energy usage for the meter located at Site 38. The missing data are handled using MICE imputation (Multiple Imputation by Chained Equation), wherein each of the missing data is replaced with m values obtained from iterating m times. The clean dataset has 49,000 rows. To deal with irregular time gaps we calculated the differences in timestamps to obtain the duration between consecutive measurements. Moreover, the difference in meter readings between consecutive timestamps gives the consumption data. Environmental factors such as temperature affects power consumption by some equipment, like heating, ventilation, and air conditioning (HVAC) systems. Fig. 1 displays the consumption pattern versus the temperature.

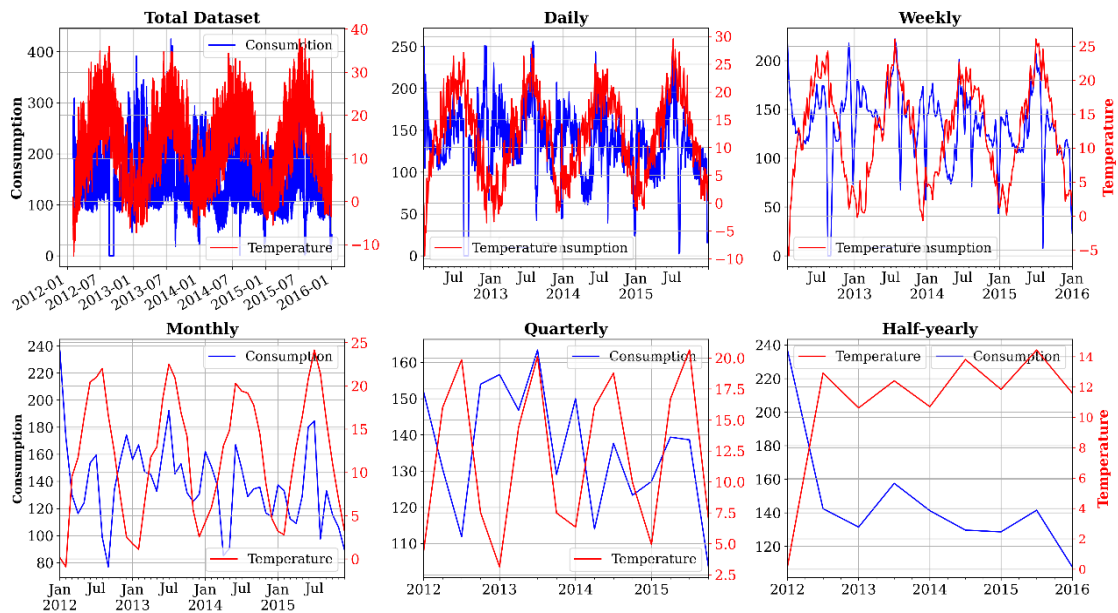

Fig. 1. Energy consumption versus temperature over time.

---

[3] https://shorturl.at/bqrM9

The data were resampled to daily, weekly, monthly, quarterly, and half-yearly prior to generating the plots, and subsequently, the mean records were taken. This provides good visibility of consumption, which is low during the early hours of the day, at its peak during the middle of the day, and slows down late in the evening. Also, we observe a higher consumption pattern during the weekdays and during the months of June and July. These are regular consumption behaviors, in line with our understanding.

Fig. 2 displays the distribution of consumption over different time intervals (days of the week, months of the year, hours of the day, and years). This helps us understand the temporal dynamics of the data.

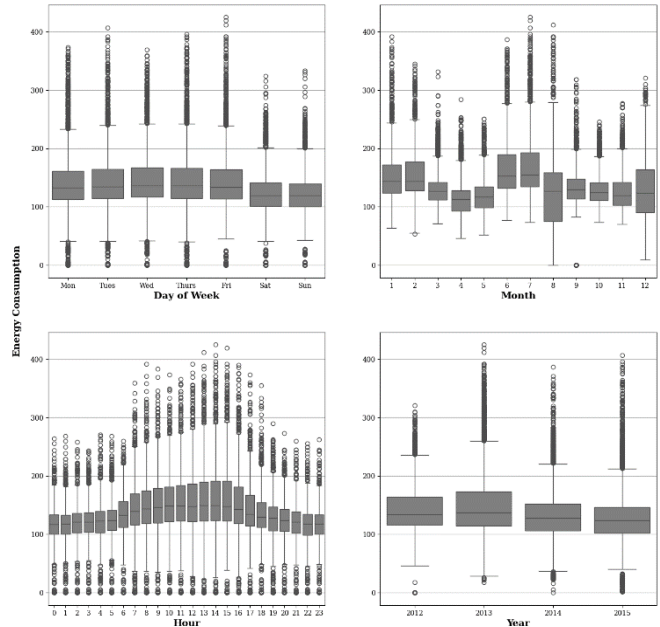

Fig 2. Consumption distribution across time intervals.

The anomalies beyond the whiskers are based on the interquartile range (IQR), which is the natural variability within the data. Though they are unusual data points, they may still be valid observations and not necessarily indicative of anomalies. Power consumption patterns often vary over time, exhibiting daily, weekly, seasonal, and even yearly trends. Here, the data is multivariate, meaning anomalies may arise from complex interactions and relationships between multiple factors, which may not be fully captured by univariate outlier detection methods.

We performed feature engineering on the explanatory variables, which include the first- and second-lag values of consumption, along with a daily and monthly lag that captures the multiple levels of seasonality. We found that just under 500 observations were not recorded every hour, which seems to be an error in data entry. However, we expect these values to still be a useful approximation of the month/weekday shift.

Fig. 3 presents the correlation matrix. We can see dayshift is significantly correlated with consumption, which makes logical sense.

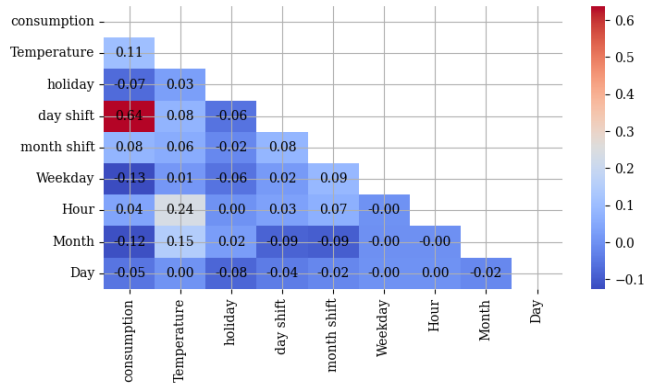

Fig. 3. Correlation plot.

Table II presents the statistical summary of the final list of columns, including the consumption (target column) and the other explanatory variables. The mean of consumption is approximately 135.68 units, with a standard deviation of 50.08 units indicating the variability in consumption levels. This is approximately 36.85% of the mean consumption value (50.08 / 135.68), indicating moderate variability. The minimum consumption value is 0 units, and the maximum is 425 units. This looks natural and can arise due to fluctuations in demand, changes in weather patterns, or periodic events.

Sudden changes in 1-lag, 2-lag consumption, day shift, and month shift (e.g., large σ, extreme minimum or maximum values) might indicate important shifts or anomalies in the data. Table III displays the statistical output. After removing the extreme values, the statistical distribution of the data looks more normal. The removal of outliers led to minor adjustments in the mean while not significantly affecting other statistical measures.

TABLE II. STATISTICAL SUMMARY.

| index | consumption | temperature | holiday | Month shift | weekday | hour | month | day |
|---|---|---|---|---|---|---|---|---|
| count | Number of data points: 33171 | | | | | | | |
| μ | 135.68 | 12.30 | 0.03 | 135.54 | 3.00 | 11.49 | 6.58 | 15.67 |
| σ | 50.08 | 8.36 | 0.16 | 49.67 | 1.99 | 6.92 | 3.36 | 8.77 |
| min | 0.00 | -12.60 | 0.00 | 0.00 | 0.00 | 0.00 | 1.00 | 1.00 |
| max | 425.00 | 37.70 | 1.00 | 425.00 | 6.00 | 23.00 | 12.00 | 31.00 |
| 25% | 110.00 | 5.70 | 0.00 | 111.00 | 1.00 | 5.00 | 4.00 | 8.00 |
| 75% | 158.00 | 18.20 | 0.00 | 160.00 | 5.00 | 17.00 | 10.00 | 23.00 |
| median | 130.00 | 12.40 | 0.00 | 132.00 | 3.00 | 11.00 | 7.00 | 16.00 |

TABLE III. STATISTICAL SUMMARY.

| index | 1-lag | 2-lags | day shift | month shift |
|---|---|---|---|---|
| count | Number of datapoints: 33167 | | | |
| $\mu$ | 135.53 | 135.69 | 135.81 | 137.54 |
| $\sigma$ | 50.06 | 50.06 | 50.04 | 49.67 |
| min | 0.00 | 0.00 | 0.00 | 0.00 |
| max | 425.00 | 425.00 | 425.00 | 425.00 |
| 25% | 110.00 | 110.00 | 110.00 | 111.00 |
| 75% | 158.00 | 158.00 | 158.00 | 160.00 |
| median | 130.00 | 130.00 | 130.00 | 132.00 |

We performed the non-parametric Kolmogorov-Smirnov goodness-of-fit test and the Anderson-Darling test. Table IV displays the statistics showing that the data is not normally distributed. The Kolmogorov-Smirnov test is more cognizant of the distribution center, while the Anderson-Darling test is far more sensitive to the distribution's tails. The analysis provides a detailed understanding of consumption patterns, identifies potential anomalies, and lays the groundwork for further anomaly detection and modeling efforts.

TABLE IV. NORMALITY TEST.

| Kolmogorov-Smirnov Test | | |
|---|---|---|
| Statistic: 0.980 | p-value: 0.0 | The data do not appear to be normally distributed |
| Anderson-Darling test | | |
| Statistic: 543.647 | Critical values: [0.576 0.656 0.787 0.918 1.092] Significance levels: [15. 10. 5. 2.5. 1] | The data do not appear to be normally distributed at the 5% significance level |

### III. METHODOLOGY

The use of Autoencoder on energy series has attracted attention from several researchers in recent times (see, [19]; [20]; etc.).

#### A. Convolutional Autoencoder

Autoencoders are effective at reconstructing data in general, especially in time-series (see [21], [22]) and researchers have reported anomaly identification in energy series using Autoencoders (see [23]; [24]; [25]). We employ a Convolutional Autoencoder that reduces noise from input data, enhancing data quality and model performance.

*1) Model training steps*

We implemented a simple network architecture consisting of three convolutional layers using batch normalization, the ReLU activation function, and max pooling. The objective is to train a model to reconstruct normal consumption patterns, detecting potential abnormalities in the data. This approach leverages both the reconstruction error and the MD to enhance anomaly detection. The model's ability to generalize is impacted by the batch size.

TABLE V. CNN-AUTOENCODER MODEL ARCHITECTURE

| Layer Type | Output Shape | Params | Activation Function | Batch Normalization |
|---|---|---|---|---|
| Input (InputLayer) | (None, 24, 1) | 0 | None | No |
| Conv1D | (None, 12, 16) | 128 | ReLU | Yes |
| BatchNormalization | (None, 12, 16) | 64 | None | Yes |
| Conv1D | (None, 6, 8) | 904 | ReLU | Yes |
| BatchNormalization | (None, 6, 8) | 32 | None | Yes |
| Conv1D | (None, 3, 4) | 228 | ReLU | No |
| Conv1DTranspose | (None, 6, 8) | 72 | ReLU | Yes |
| BatchNormalization | (None, 6, 8) | 32 | None | Yes |
| Conv1DTranspose | (None, 12, 16) | 912 | ReLU | Yes |
| BatchNormalization | (None, 12, 16) | 64 | None | Yes |
| Conv1DTranspose | (None, 24, 1) | 113 | None | No |

Moreover, smaller batch sizes introduce more randomness, which helps escape local minima and leads to better exploration of the loss landscape. We experimented with commonly used batch sizes, such as powers of 2 (e.g., 16, 32, 64, 128) prior to fine-tuning the batch size (64) based on empirical results and observations during training. Fig. 5 provides insights into how well the model is learning to reconstruct the input sequences over the course of training epochs.

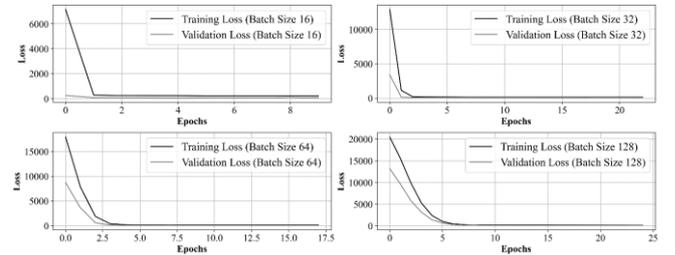

Fig. 4. Learning curves of the Autoencoder model [Source: Authors].

Smaller batch sizes introduce more noise into the parameter updates, which improves the model's ability to generalize to new data. We can see that both the training and validation curves decrease steadily during the early epochs of training. This indicates that the model is learning to reconstruct the input sequences effectively. Table V displays an overview of the architecture, which consists of an encoder followed by a decoder. The encoder compresses the input data into a latent space representation, and the decoder reconstructs the original input from this.

Fig. 6 displays the distribution of reconstruction errors. The distribution is concentrated towards the left side of the plot, indicating most input sequences are well-reconstructed by the Autoencoder. We further employ a moving average to smooth noisy data, identify trends, establish a baseline, and calculate a dynamic threshold. This threshold adjusts to changes in data distribution over time, making the process more adaptive and responsive to data variations. It also helps identify potential anomalies by detecting sudden deviations from the moving average.

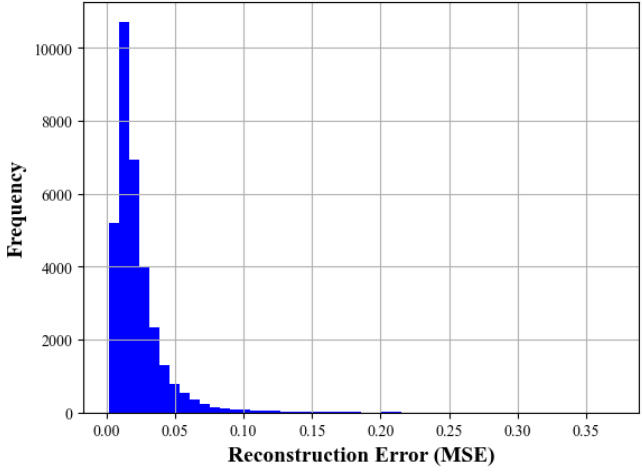

Fig. 5    Distribution of Reconstruction Errors.

To specify the exact time at which anomalies occur, we introduced a novel method that integrates the output of the Autoencoder with the MD. The MD is calculated based on selected features such as temperature, holiday, month shift, weekday, hour, month, and day. This provides a measure of the distance of each data point from the expected distribution. Equations 1–7 explain how the unusual patterns in the data are identified. By combining MSE (reconstruction errors) with MD (considering the distribution and correlation of features), the algorithm can better distinguish anomalies from normal data points. The use of moving averages and dynamic thresholds adds adaptability to changing patterns in the data over time.

$$MSE = \frac{1}{n}\sum_{i=1}^{N}(x_i - x'_i)^2 \quad (1)$$

$$MD = \sqrt{(x-\mu)^T \Sigma^{-1}(x-\mu)} \quad (2)$$

$$CombinedScore\ (CS) = MSE + MD \quad (3)$$

$$MovingAverage\ (MV) = \frac{1}{w}\sum_{i=1}^{N-W+1} CS_i \quad (4)$$

$$MV_{std} = \sqrt{\frac{1}{N\ w}\sum_{i=1}^{N-W+1}(CS_i - MA)^2} \quad (5)$$

$$DynamicThreshold = MA + k * MV_{std} \quad (6)$$

$$Anomaly = \begin{cases} 1, & if\ CS > DynamicThreshold \\ 0, & otherwise \end{cases} \quad (7)$$

where $N$ = number of datapoints, $x_i$ and $x'_i$ are the $i^{th}$ data points in the original and reconstructed sequences respectively, $x$ is a datapoint in the dataset, $\mu$ is the mean vector of data, $\Sigma$ is the covariance matrix of data. W = window size, $CS_i$ = combined score at time i, and multiplier k is used to adjust the sensitivity of the threshold.

The model has detected 622 anomalies from the unseen test dataset (Fig. 7). Fig. 8 is an extension of Fig. 7 with the timestamp, where the red dots are the anomalous points with the timestamp.

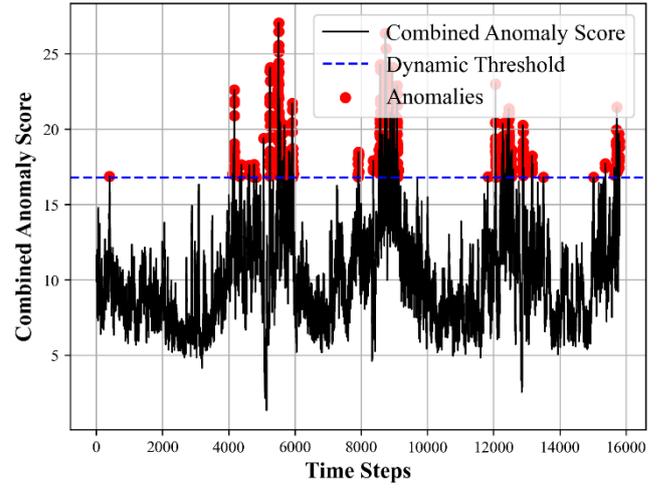

Fig. 6.    Combined anomaly score with dynamic threshold.

The proposed approach is close to a real-life deployment, which makes it well-suited for deployment in real-life energy monitoring systems.

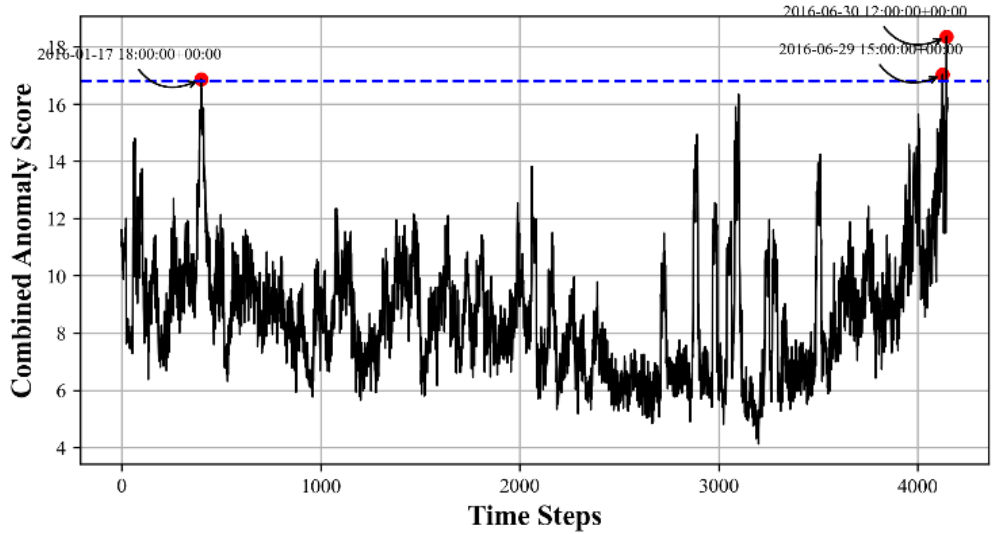

Fig. 7.    Anomalies with timestamp.

The approach is interpretable and displays a timestamp for immediate human inference. Visualization is an effective method here to immediately discover probable anomalies in the data.

## IV. LIMITATIONS AND FUTURE DIRECTIONS

The stability and accuracy of consumption data are the basis of anomaly detection here. Thus, inconsistencies, missing values, or noise in the data can affect the accuracy model. Moreover, the proposed system combines statistical methods with CNN. Thus, the interpretability of the detected anomalies may be challenging. Understanding the exact cause of anomalies and taking appropriate actions based on them may require further investigation and domain expertise. The dynamic thresholds based on moving averages and MD may require fine-tuning and adjustment for different datasets. Selecting an appropriate threshold that balances false positives and false negatives can be challenging. These limitations would require further research and refinement of the proposed methodology, potentially exploring hybrid approaches combining different ML techniques or leveraging additional domain knowledge.

## V. CONCLUSION

This study presents a dynamic anomaly detection framework that integrates Convolutional Neural Networks (CNN) with Autoencoder models. The approach captures spatial patterns and correlations in consumption series data, enhancing anomaly detection accuracy. The model uses reconstruction error and Mahalanobis distance to understand data patterns and anomalies. Moving averages smooth noisy data and establish dynamic thresholds, enabling real-time anomaly detection. The framework successfully detected 622 anomalies in unseen test data, providing valuable insights for power consumption monitoring and management.